%% file: iclr2021_conference.tex
\documentclass{article} 
\usepackage{iclr2021_conference,times}

\input{math_commands.tex}

\usepackage{hyperref}
\usepackage{url}
\usepackage{graphicx}

\title{Remote Diffusion}

\author{Kunal Sunil Kasodekar \thanks{Collaborator - Kartik Jawanjal (kjawnja@asu.edu)
} \\
SCAI\\
Arizona State University\\
Tempe, AZ 85281, USA \\
\texttt{kkasodek@asu.edu} \\
}

\iclrfinalcopy 
\begin{document}

\maketitle

\begin{abstract}

I explored adapting Stable Diffusion v1.5 for generating domain-specific satellite and aerial images in remote sensing. Recognizing the limitations of existing models like Midjourney and Stable Diffusion, trained primarily on natural RGB images and lacking context for remote sensing, I used the RSICD dataset to train a Stable Diffusion model with a loss of 0.2. I incorporated descriptive captions from the dataset for text-conditioning. Additionally, I created a synthetic dataset for a Land Use Land Classification (LULC) task, employing prompting techniques with RAG and ChatGPT and fine-tuning a specialized remote sensing LLM. However, I faced challenges with prompt quality and model performance. I trained a classification model (ResNet18) on the synthetic dataset achieving 49.48\% test accuracy in TorchGeo to create a baseline. Quantitative evaluation through FID scores and qualitative feedback from domain experts assessed the realism and quality of the generated images and dataset. Despite extensive fine-tuning and dataset iterations, results indicated subpar image quality and realism, as indicated by high FID scores and domain-expert evaluation. These findings call attention to the potential of diffusion models in remote sensing while highlighting significant challenges related to insufficient pretraining data and computational resources.

\end{abstract}

\section{Introduction}

Image generation models are the talk of the town due to rapid development in the field of Diffusion Models. Ever since the advent of Midjourney, Stable Diffusion \cite{rombach2022highresolution}, and SORA, diffusion models have replaced GANs for generating realistic, robust, and high-quality images with text conditioning. However, these generative models are trained to generate natural images (RGB), which differ from multi-spectral images used in the remote sensing domain. Hence, these models cannot be used for generating satellite and aerial images for the remote sensing domain as they lack specific domain knowledge and context. 
Satellite and aerial images play a crucial role in remote sensing, aiding in addressing various challenges such as food security through precision farming, disaster forecasting, and the detection of wildfires and deforestation \cite{neumann2019domain}. Therefore, it is highly advantageous to produce realistic, high-quality satellite and aerial images for any scenario or downstream task, whether for augmentation, simulation purposes or when generating such images is not feasible \cite{sebaq2023rsdiff}. In any case, it presents an interesting and fun problem to tackle.

Only two works \cite{sebaq2023rsdiff, khanna2023diffusionsat} regarding domain-specific remote sensing diffusion models have been published at the time of writing this report. Therefore, to fill this gap, I trained a Stable Diffusion model on the RSICD dataset to create a domain-specific generative model. Furthermore, the dataset provides accompanying descriptive sentences that I have used for text-conditioning the model during its training. I have conducted a quantitative analysis by calculating the FID score, which gives the measure of similarity between ground truth and generated images. For qualitative analysis, the opinion of a domain expert is taken.
Furthermore, compared to previous works, I curated a new out-of-domain synthetic dataset for a Land Use Land Classification (LULC) downstream task. I generated a dataset of 388 images for 7 LULC classes, wherein the images are conditioned on text prompts generated through a combination of RAG (Retrieval Augmented Generation) and ChatGPT. I fine-tuned a Resnet18-RGB-Moco \cite{wang2023ssl4eos12} pretrained on the Sentinel-2 RGB dataset for the above downstream task. I created a Llama-Index-based RAG pipeline that uses Mistral-7b for the LLM, chroma db for the vector db, and the aforementioned corpus for augmentation. Additionally, I also fine-tuned a remote sensing-specific LLM (Microsoft-Phi 1.5 \cite{li2023textbooks}) on a text corpus curated from research papers presented in this class. However, despite trying out multiple prompting techniques, the prompts generated are subpar and not suitable for conditioning SDv1.5. \textit{Nevertheless, the model itself can understand remote sensing-specific jargon and provide generic answers}.


\section{Related Work}
\cite{sebaq2023rsdiff} proposed RSDiff, a two-stage diffusion model conditioned on text. Despite its superiority over existing methods on the RSICD dataset, RSDiff struggled with generating high-quality images and incurred computational expenses due to incomplete latent diffusion. \cite{khanna2023diffusionsat} introduced DiffusionSat, a remote-sensing latent diffusion model inspired by Stable Diffusion. While it contributed to foundational models in remote sensing, it focused more on implementation details than downstream applications and synthetic dataset curation. \cite{10327767} presented a method for generating diverse hyperspectral remote sensing images (HSI) using RGB images as conditional inputs. They utilized HyperLDM for stable and noise-reduced synthesis, along with conditional vector quantization VQGAN for generating latent vectors. However, further exploration of its applicability across diverse geographic domains and exploration of diffusion techniques is deserved.

\section{Dataset}
\label{headings}
I utilized the RSICD (Remote Sensing Image Captioning Dataset) \cite{lu2017exploring} for training my Stable Diffusion Model, SD version - 1.5. This dataset was originally employed for an image captioning task as outlined in the RSICD paper. The images are sourced from Google Earth, Baidu Map, and MapABC, and are sized at 224x224 pixels with varying resolutions, displaying high intra-class diversity and low inter-class diversity. It comprises 30 classes and 10,921 images, with an uneven distribution across the classes. Each image is described through five sentences, annotated by multiple volunteers with domain knowledge and annotation experience. The dataset is converted into a format compatible with the diffusers SD training pipeline. Specifically, the dataset, which is initially in parquet format, needs to be converted into an image batch file structure and a metadata JSON file compliant with SDv1.5. For the train-test split, I reserved 500 images for conducting quantitative analysis later, including calculating the FID score. Regarding preprocessing, I resized all the images to an input resolution of 224x224 pixels and applied all seven possible flip and rotate transforms as a form of data augmentation. Additionally, I normalized the dataset by subtracting the mean of the entire dataset and dividing it by the standard deviation of the entire dataset.

Regarding the synthetic dataset, for the LULC downstream task, the generated 388 images are sized at 512x512 pixels with 7 class labels. See Table~\ref{LULC Dataset Details}. To produce clean, realistic images, I employed a PNDMscheduler with 50 inference steps, utilizing positive prompt cues to facilitate the generation of realistic satellite/aerial images and negative cues to discourage \textit{wrapped, repeating, blurry, deformed, and low-quality} images. While it is feasible to curate a larger dataset, for my experiment and considering current computational limitations, this size should suffice for the downstream task. The created dataset has been publicly uploaded to Hugging Face in a parquet format, available for open-source consumption.

\section{Methodology}
\label{gen_inst}

Initial attempts to train the entire model from scratch were futile due to a lack of computational resources and the absence of a large-scale dataset with image-caption pairs. While it was possible to create a large dataset from scratch, the absence of a remote sensing-specific captioning model and limited time to curate such a pipeline and dataset prevented this from happening. Furthermore, training a Stable Diffusion model from scratch to achieve visually appealing results typically requires weeks, even with a cluster of A100 GPUs. Hence, I opted to fine-tune a Runway-Stable Diffusion v1.5 model to build upon semantic understanding and accurate generation, aiming for more nuanced, precise, and contextually relevant outputs.

\begin{figure}[h]
\begin{center}
\includegraphics[width=0.45\textwidth]{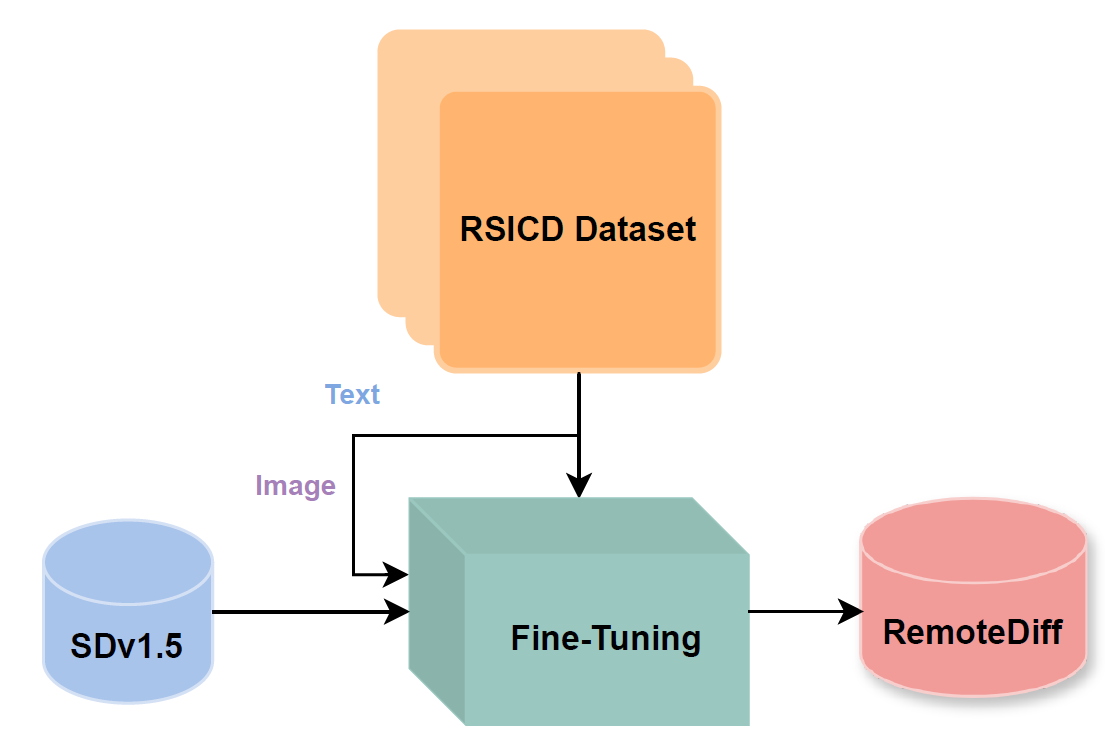}
\end{center}
\caption{Fine-Tuning}
\label{i1}
\end{figure}

I utilized the Diffusers library from Hugging Face to fine-tune Stable Diffusion v1.5 from RunwayML \cite{Rombach_2022_CVPR}. See Figure~\ref{i1}. I conducted multiple fine-tuning runs to achieve the best possible results. The training process is checkpointed every 500 steps, and at the 2500th step, a minimum loss of \textbf{0.2} is observed. According to the training documentation, intermediate checkpoints may yield better results, and in this case, the 2500th step demonstrates the best result, as confirmed through WandB experiment tracking. Fine-tuning for more than 5 epochs resulted in poor performance in my case, as the model began to forget its inherent understanding and generation capabilities of natural images, leaning more towards overfitting satellite/aerial images. This led to a loss of its ability to generate diverse, realistic, and novel images, instead producing more averaged-out satellite images from the dataset. Further training only resulted in average images. Therefore, it is concluded that training for more epochs is only feasible when the dataset is sufficiently large compared to the original pretraining dataset LAION-5b. The model was trained for 5 epochs with a batch size of 4, a learning rate of 1e-06, in mixed precision mode, with 4 gradient accumulation steps.

\begin{figure}[h]
\begin{center}
\includegraphics[width=0.65\textwidth]{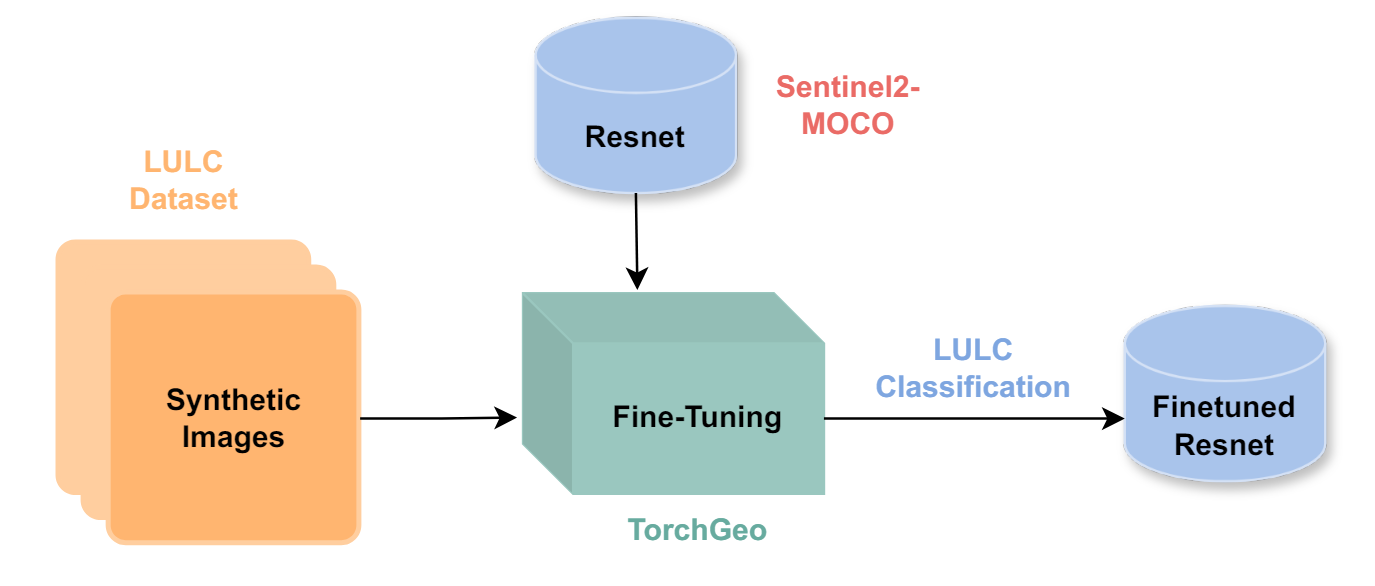}
\end{center}
\caption{Downstream Evaluation}
\label{i3}
\end{figure}

Once the dataset is created for downstream evaluation according to the process outlined in the dataset section, I loaded the Parquet dataset from Hugging Face using the datasets module. I transformed the dataset with Albumentations, including random cropping, image normalization, flips, and random brightness contrast. The Parquet dataset is then converted to a PyTorch Lightning DataModule through a custom \textit{pl.LightningDataModule}. 
For training, I instantiated a TorchGeo \cite{stewart2022torchgeo} classification task with a ResNet-18 model, utilizing Sentinel2-RGB-Moco weights, and a trainer that runs for 20 epochs (determined as the best number based on multiple experiments) to train the LULC classification model. See Figure~\ref{i3}.

\begin{figure}[h]
\begin{center}
\includegraphics[width=0.75\textwidth]{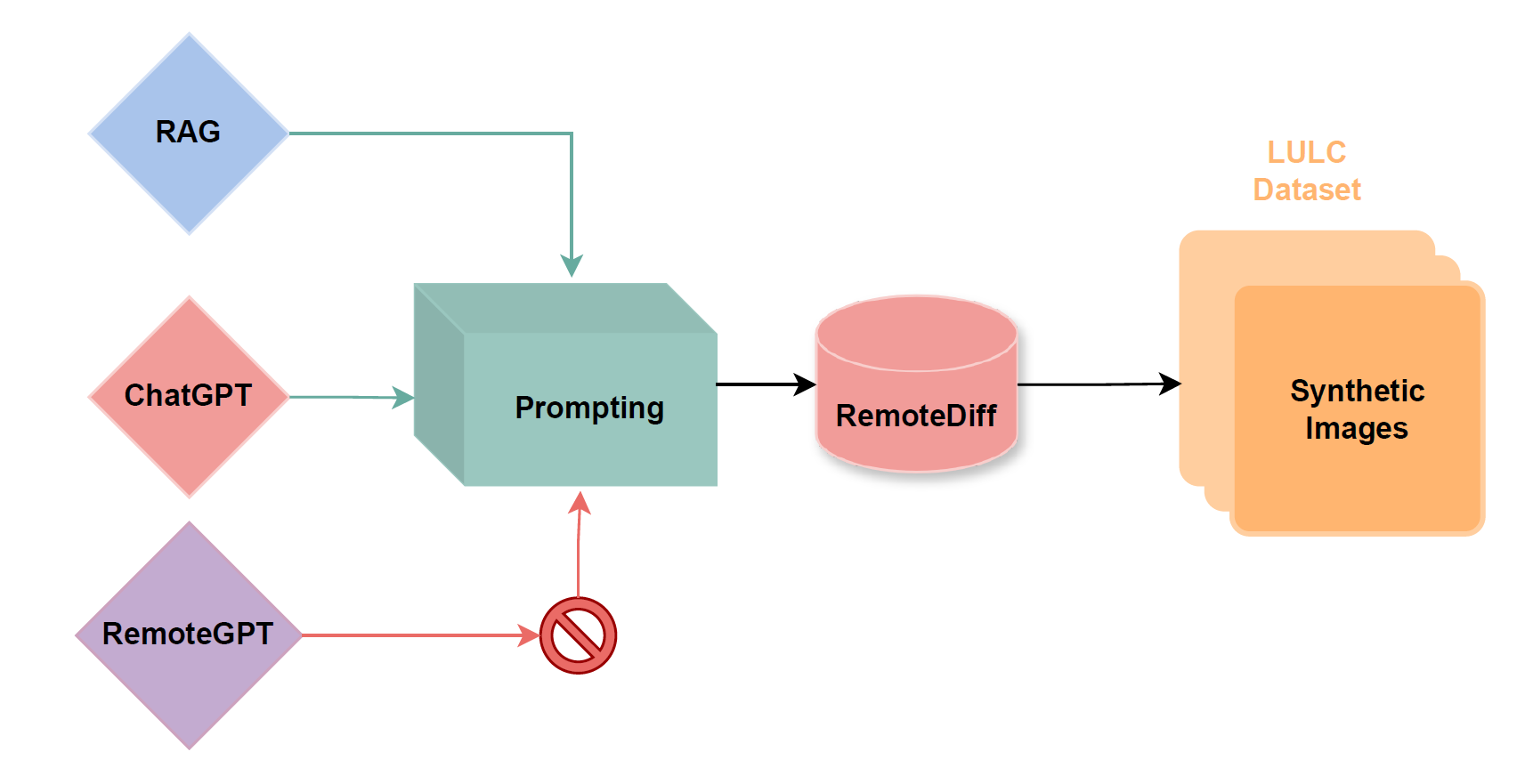}
\end{center}
\caption{Data Generation}
\label{i2}
\end{figure}

To create the remote-sensing text corpus, I scraped text from 50 small to medium-sized relevant ML for Remote Sensing research papers using PyPDF. This concatenated text is converted to a Parquet dataset with 71,102 rows, with each row averaging around 83 characters in length, with 5\% of the dataset set aside as a test set. The dataset is then broken down into chunks of at least 500 characters until the next nearest full stop, and an End-of-Sentence (EOS) token is added at the end of each chunk to signal the model to stop generation. 
This dataset is tokenized with a context length of 512 using the Phi-1.5 tokenizer trained using the Hugging Face causal LLM library with a QLoRa config (PEFT Library \cite{xu2023parameterefficient}) of lora\_alpha = 8, rank = 16, and target modules as k, q, v in the self-attention operation, along with a dropout of 0.05. QLoRa reduces the GPU memory by 10 fold and the training parameters by 1000 fold along with quantization. The model is trained with a learning rate of 2e-5 for 5 epochs with a weight decay of 0.01, using a data collator that appends the EOS token for padding.
For RAG augmentation, the aforementioned corpus is loaded with a chunk size of 256 and indexed using a Chroma db vector store. Llama\_index with the Mistral-7b \cite{jiang2023mistral}model is used as the chaining library and LLM, respectively. I queried the index for augmenting and grounding the prompts generated by the LLM. See Figure~\ref{i2}.

\section{Experiments}

For the downstream task, I curated a synthetic dataset comprising 388 images of size 512x512 pixels, categorized into 7 classes for Land Use and Land Cover (LULC) classification. For the baseline, I developed a classification task in TorchGeo utilizing a ResNet-18 model pretrained on Sentinel-2 RGB-MOCO. I fine-tuned this model on the aforementioned dataset, using a learning rate of 3e-4 for 20 epochs with cross-entropy loss, resulting in a test accuracy of \textbf{49.48 percent}. See Table~\ref{Downstream Evaluation}.
For the remote-sensing specific LLM, I fine-tuned the Microsoft Phi-1.5 on the remote-sensing-corpus mentioned in the dataset section for 3 epochs with a learning rate of 2e-5, weight decay set to 0.01, using a batch size of 8, and for 3000 steps. The model achieves a minimum training loss of \textbf{3.3855} at the 2748th step and a \textbf{perplexity score of 30.5} with a QLoRa training configuration and a causal LLM trainer from Hugging Face.
To conduct quantitative analysis, the FID score is calculated to assess the quality of the images generated by the Diffusion Model. This metric compares the distribution of the ground truth images from the test set with the distribution of the images generated by the diffusion model using the same ground truth captions. A lower FID score implies that the generated images more closely resemble real ground truth images. Light preprocessing is applied to normalize the images, and then both the generated and ground truth images are stacked in a NumPy array. I utilized FID from Torchmetrics and, after multiple sampled runs of 250 images, found an average FID score of \textbf{245.3629}, which is considered objectively undesirable as it is on the higher side. 
For a qualitative analysis, I sought the expertise of Earth Data specialist, Prof. Hannah Kerner. For more details on the task and metrics see Table~\ref{Evaluation Metrics}.



\begin{table}[t]
\begin{minipage}{0.5\textwidth}
\begin{center}
\caption{Downstream Evaluation}
\vspace{0.5em}
\label{Downstream Evaluation} 
\begin{tabular}{ll}
\multicolumn{1}{c}{\bf Metric} & \multicolumn{1}{c}{\bf Value} \\
\\ \hline \\
Test Loss & 1.555 \\
Test Average Accuracy & 0.495 \\
Test F1 Score & 0.436 \\
Test Jaccard Index & 0.280 \\
Test Overall Accuracy & 0.436 \\
\end{tabular}
\end{center}
\end{minipage}%
\begin{minipage}{0.5\textwidth}
\begin{center}
\caption{LULC Dataset Details} 
\vspace{0.5em}
\label{LULC Dataset Details}
\begin{tabular}{ll}
\multicolumn{1}{c}{\bf Classes} & \multicolumn{1}{c}{\bf Images} \\
\\ \hline \\
Bare Land & 52 \\
Crop Land & 57 \\
Cultivated Vegetation & 54 \\
Natural Vegetation & 54 \\
Snow Ice & 58 \\
Water Body & 52 \\
Woody Vegetation & 61 \\
\end{tabular}
\end{center}
\end{minipage}
\end{table}

\section{Discussion and Conclusion}

The SD model can generate domain-specific, high-resolution images using positive and negative prompts. Prompting offers additional control to generate either \textit{high} or \textit{low-resolution} satellite/aerial images. While the generated images initially appear realistic and of high quality, a closer inspection reveals \textit{flaws}. According to the evaluation by domain experts, although the images initially seem realistic, they suffer from low quality due to artifacts, repetitions, or a lack of diversity in the overall image structure. In essence, thorough observation reveals a lack of expected realism in the images. The main reason for this performance is a lack of data during pretraining and insufficient computational resources for experimentation. 
Despite training a domain-specific LLM to streamline prompt generation, it \textit{failed} to generate well-formed prompts for class-specific data. RAG and ChatGPT were utilized instead, as they outperformed the LLM.
The FID evaluation yielded a score of 245.3629, which is relatively high. However, Stable diffusion's output heavily depends on the text prompt provided and may vary based on the number of sampling steps, even for the same prompt. Despite potential differences in the satellite image scene, semantically, it \textit{fulfils the requirements} stated by the text prompt. The domain-specific LLM did not meet expectations, even after utilizing advanced prompting techniques for prompt generation. This could be attributed to the small size of the corpus, which was insufficient for fine-tuning compared to other state-of-the-art text datasets. 
Due to limited computational resources, a smaller causal LLM (Microsoft Phi-1.5) was used, resulting in shorter training epochs. With increased computational resources and a larger corpus, better domain-specific outputs could be achieved. While images generated for classes like Bare Land, Snow/Ice, and Water Body appeared realistic to non-domain experts, others like Cultivated Vegetation, Woody Vegetation, Natural Vegetation, and Crop Land showed poor quality due to low inter-class diversity, leading to semantic inaccuracies, image repetitions, and model bias. For example, using the prompt \textbf{a pumpkin field} for the Crop Land class produced an image with large pumpkins, indicating bias.

In conclusion, my work is among the first to address the gap in domain-specific remote sensing diffusion-based text-to-image generation models. I trained a Stable Diffusion v1.5 model on the RSICD dataset and achieved a minimum loss of \textbf{0.2}. Furthermore, it is the first attempt to create a synthetic dataset for an out-of-domain downstream task, and class and establish a baseline. Despite multiple efforts to fine-tune the model and create multiple versions of the dataset, the results indicate suboptimal performance in terms of image quality and realism, specifically for satellite/aerial images. The high FID score and domain expert evaluation conclude that the generated images lack the desired quality, diversity, and realism required for remote sensing data, highlighting the challenges in adopting diffusion models for the remote sensing domain. The major reason for this performance is a lack of data during pretraining and a lack of compute for experimentation. To streamline the process of prompt generation, I tried to train a domain-specific LLM; however, despite advanced prompting techniques, it doesn't generate well-formed prompts for generating class-specific data for downstream tasks. The model trained on the downstream task achieves a low accuracy, as some of the classes have very little inter-class diversity, making it difficult for the diffusion model to generate diverse images and for the downstream model to create efficient decision boundaries. 
One limitation is the absence of location and time embeddings, crucial for truly remote sensing-based data, primarily because RSICD lacks time and location metadata. To address this, Geo-Clip or datasets with such metadata could be utilized. However, incorporating time and location embeddings typically necessitates training the entire model from scratch, posing challenges due to limited data and computational resources, but future advancements in both could resolve this. In the future, I plan to train a BLIP model on remote-sensing data and caption large earth datasets to address data scarcity, enabling the creation of \textbf{robust, diverse, and realistic} synthetic datasets for remote-sensing tasks.


\bibliography{iclr2021_conference}
\bibliographystyle{iclr2021_conference}

\appendix
\section{Appendix}

\subsection{Model and Dataset Links}
\textbf{RemoteDiffusion:}\url{https://huggingface.co/gremlin97/RemoteDiff}

\textbf{RemoteGPT:}\url{https://huggingface.co/gremlin97/remote_sensing_gpt_expt4}

\textbf{LULC-Synthetic-Dataset:}\url{https://huggingface.co/datasets/gremlin97/LULC-Synthetic-Dataset}

\textbf{Remote-Text-Corpus:}\url{https://huggingface.co/datasets/gremlin97/RemoteSensingCorpus}

\subsection{Synthetic Dataset Curated Sample Images}




\begin{figure}[h]
\centering
\begin{minipage}{.31\textwidth}
  \centering
  \includegraphics[width=\linewidth]{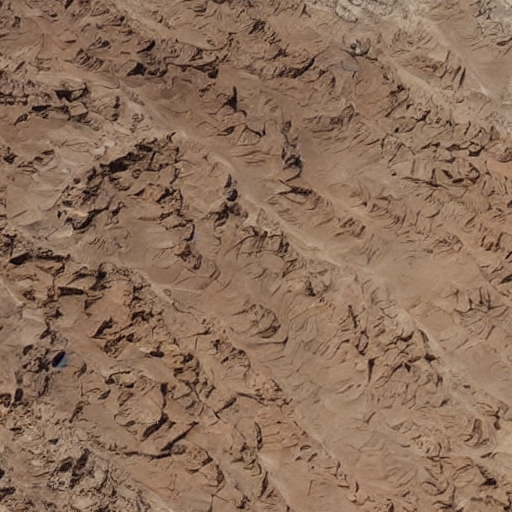}
  \caption{Prompt: Barren landscape of a rocky desert canyon, Class: Bare Land}
\end{minipage}%
\hfill
\begin{minipage}{.3\textwidth}
  \centering
  \includegraphics[width=\linewidth]{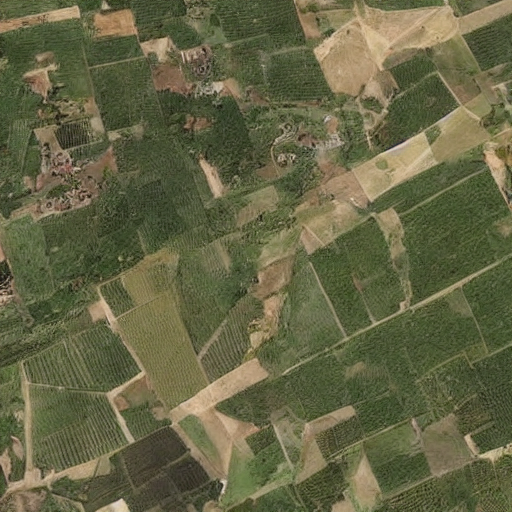}
  \caption{Prompt: Vineyards and orchards in a wine-producing region, Class: Crop Land}
\end{minipage}
\hfill
\begin{minipage}{.3\textwidth}
  \centering
  \includegraphics[width=\linewidth]{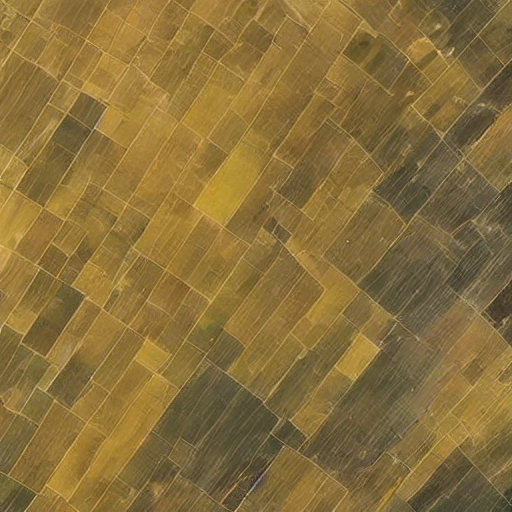}
  \caption{Prompt: Golden hues of ripe wheat fields ready for harvest, Class: Cultivated Vegetation}
\end{minipage}
\end{figure}

\begin{figure}[h]
\centering
\begin{minipage}{.45\textwidth}
  \centering
  \includegraphics[width=\linewidth]{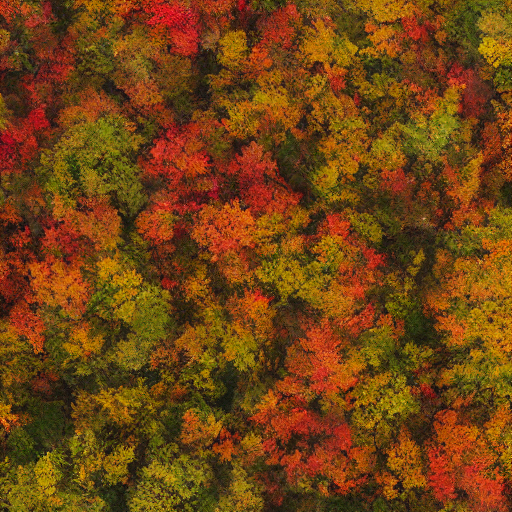}
  \caption{Prompt: Mixed oak-hickory forest with vibrant autumn foliage, Class: Natural Vegetation}
\end{minipage}%
\hfill
\begin{minipage}{.45\textwidth}
  \centering
  \includegraphics[width=\linewidth]{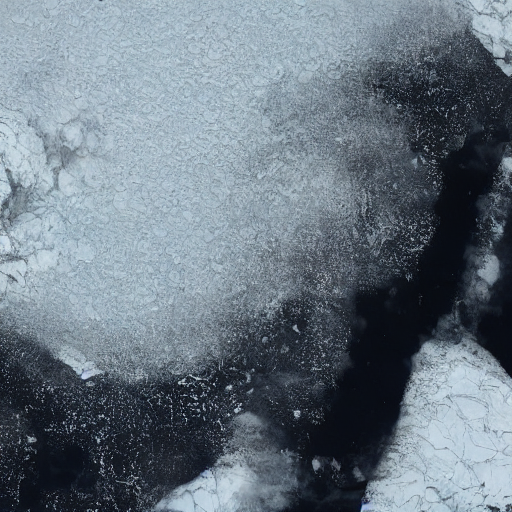}
  \caption{Prompt: Ice floes drifting in the Arctic Ocean under the northern lights, Class: Snow/Ice}
\end{minipage}
\end{figure}

\begin{figure}[h]
\centering
\begin{minipage}{.45\textwidth}
  \centering
  \includegraphics[width=\linewidth]{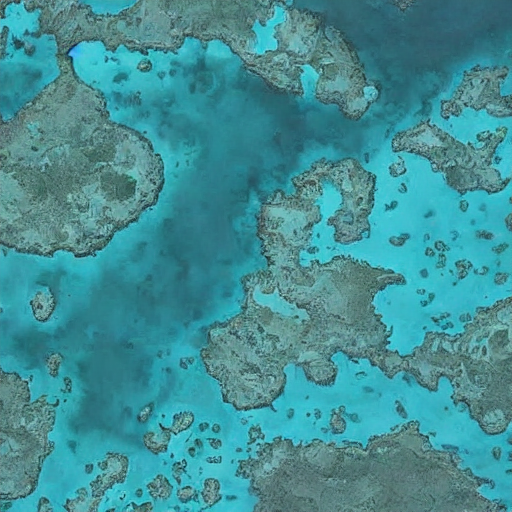}
  \caption{Prompt: Cluster of small islands surrounded by shallow turquoise waters, Class: Water Body}
\end{minipage}%
\hfill
\begin{minipage}{.45\textwidth}
  \centering
  \includegraphics[width=\linewidth]{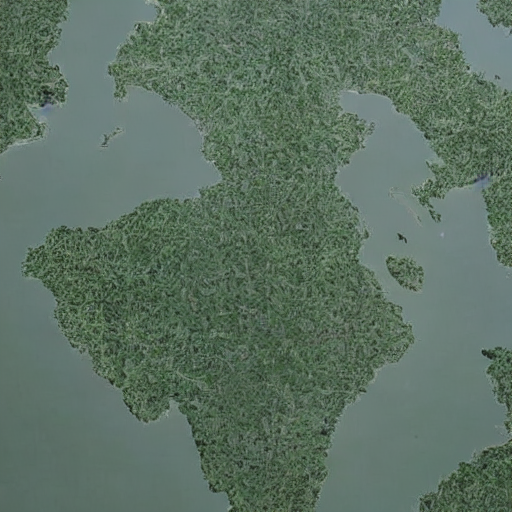}
  \caption{Prompt: Coastal mangrove swamp with meandering tidal creeks, Class: Woody Vegetation}
\end{minipage}
\end{figure}

\subsection{Evaluation Metrics Table}

For task-wise evaluation metrics see Table~\ref{Evaluation Metrics}.

\begin{table}[t]
\caption{Metrics Table}
\vspace{0.5em}
\label{Evaluation Metrics}
\begin{center}
\begin{tabular}{lll}
\multicolumn{1}{c}{\bf Task}  &\multicolumn{1}{c}{\bf Metric} &\multicolumn{1}{c}{\bf Score}
\\ \hline \\
Quantitative Evaluation & FID & 245.3629 \\
Stable Diffusion Training & Sinkhorn Divergence & 0.2 \\
LLM Finetuning & Perplexity loss & 30.5 \\
\end{tabular}
\end{center}
\end{table}

\end{document}

%% file: math_commands.tex
\usepackage{amsmath,amsfonts,bm}

\def\eqref#1{equation~\ref{#1}}

\def\1{\bm{1}}

\DeclareMathAlphabet{\mathsfit}{\encodingdefault}{\sfdefault}{m}{sl}
\SetMathAlphabet{\mathsfit}{bold}{\encodingdefault}{\sfdefault}{bx}{n}

